\documentclass{article}

\usepackage{PRIMEarxiv}

\usepackage[utf8]{inputenc} % allow utf-8 input
\usepackage[T1]{fontenc}    % use 8-bit T1 fonts
\usepackage{hyperref}       % hyperlinks
\usepackage{url}            % simple URL typesetting
\usepackage{booktabs}       % professional-quality tables
\usepackage{amsfonts}       % blackboard math symbols
\usepackage{nicefrac}       % compact symbols for 1/2, etc.
\usepackage{microtype}      % microtypography
\usepackage{lipsum}
\usepackage{fancyhdr}       % header
\usepackage{graphicx}       % graphics
\graphicspath{{media/}}     % organize your images and other figures under media/ folder
\usepackage{amsmath}
\usepackage[url=false]{biblatex}
\newcommand{\pop}[1]{{}}

\title{LaboRecommender: A crazy-easy to use Python-based recommender system for laboratory tests
}

\author{
  Fabián Villena \\
  www.fabianvillena.cl \\
  Chile \\
  \texttt{villenafabian@gmail.com} \\
}

\begin{document}
\maketitle

\begin{abstract}
Laboratory tests play a major role in clinical decision making because they are essential for the confirmation of diagnostics suspicions and influence medical decisions. The number of different laboratory tests available to physicians in our age has been expanding very rapidly due to the rapid advances in laboratory technology. To find the correct desired tests within this expanding plethora of elements, the Health Information System must provide a powerful search engine and the practitioner need to remember the exact name of the laboratory test to correctly select the bag of tests to order. Recommender systems are platforms which suggest appropriate items to a user after learning the users’ behaviour. A neighbourhood-based collaborative filtering method was used to model the recommender system, where similar bags, clustered using nearest neighbours algorithm, are used to make recommendations of tests for each other similar bag of laboratory tests. The recommender system developed in this paper achieved 95.54 \% in the mean average precision metric. A fully documented Python package named LaboRecommender was developed to implement the algorithm proposed in this paper.
\end{abstract}

% keywords can be removed

\section{Introduction}
\pop{Importance of laboratory tests in clinical decision making.}Laboratory tests (we are going to be referring to clinical chemistry tests specifically) play a major role in the clinical decision making, because they are essential for the confirmation of diagnostics suspicions and influence medical decisions in general. A large proportion of clinical encounters, inpatient and outpatient, requires laboratory testing, henceforth, the value of these diagnostic procedures is significant. Literature dictates that overall 35 \% of the clinical encounters requires at least one laboratory test and for inpatient the proportion goes up to 98 \% \cite{ngo_frequency_2017}.

\pop{Large number of laboratory tests difficult an efficient search.}The number of different laboratory tests available to physicians in our age has been expanding very rapidly to nearly 3000 different laboratory tests, this number will continue to increase due to the rapid advances in laboratory technology \cite{wians_clinical_nodate}. To find the correct desired tests within this expanding plethora of elements, the Health Information System (HIS) must provide a powerful search engine and the practitioner need to remember the exact name of the laboratory test to correctly select the bag of tests to order. 

The process of selection of the laboratory tests, taking into account the attributes described above, is time-consuming and prone to errors by misselection or omission in the clinical order entry process \cite{zhi_landscape_2013}.

\pop{Normally, groups of laboratory tests are requested at the same time, mainly because clinical guidelines.}Typically laboratory tests are ordered through a logic of \textit{order sets}, this term refers to a selection of a group of laboratory tests related to each other to answer a specific clinical question. These clinical questions are often operationalized in clinical guidelines based on the state of the art evidence to describes the best route to manage certain clinical events. Therefore is common and suggested to order laboratory tests in bags \cite{chan_order_2012}.

\pop{An automatic recommendation of tests based on already selected tests could improve efficiency and accuracy in requests.}Based on the premise of the existence of laboratory test bags there is an intuition for solving the problem of the resource-intensive selection of laboratory tests by filling the bags semiautomatically by suggesting to the practitioners the most probable tests to select next based on currently present tests in the bag. This clinical decision support (CDS) solution I am proposing is called a \textit{recommender system}. There is evidence dictating that CDS systems applied to laboratory test orders improve the quality of the orders, clinical outcomes and cost-effectiveness \cite{delvaux_effects_2017,cadogan_effectiveness_2015}.

% \pop{Data science and learning from historical data}For the determination of the so-called laboratory test bags and the modelling of the recommender system, an unsupervised-learning approach is going to be used over historical data of laboratory tests orders. This approach omits the stage of manual determination of the bags and the set of bags is going to be computed and represented automatically through data science techniques for the modelling of the recommender system.

\pop{A python package is developed}Besides the construction, training and validation of the recommender system for laboratory tests, an easy to use \texttt{Python} package is developed, documented and released to the public\footnote{LaboRecommender -  \href{https://github.com/fvillena/laborecommender}{https://github.com/fvillena/laborecommender}}, implementing the best findings of this paper.

% \subsection{Recommender systems and collaborative filtering}

\pop{Recommender systems definition.}Recommender systems are platforms which suggest appropriate items to a user after learning the users' behaviour. In this work, I want to recommend a laboratory test addition to a bag of laboratory tests to a practitioner, based on already added laboratory tests. These systems use information filtering to recommend information of interest to a user \cite{alyari_recommender_2018}.

\pop{Recommender systems in medicine.}Recommender systems are a technology that has been used in a variety of fields, ranging from social media to healthcare. Specifically in medical informatics, these systems have been developed to support clinical decisions, for example, recommender systems are used in food recommendations for diabetic patients \cite{norouzi_mobile_2018}, suggesting cardiac diagnostics \cite{mustaqeem_modular_2020} and suggesting where to publish medical papers \cite{feng_deep_2019}.

\pop{Collaborative filtering.}The basic principle of recommender systems is that significant dependencies exist between users and items. These dependencies can be modelled in a data-driven manner through historical interaction data. Many different approaches can be used to model this process and the most used method is \textit{collaborative filtering}, which refers to the use of interaction from multiple users to predict future interactions of similar users. In this work, a neighbourhood-based collaborative filtering method was used \cite{aggarwal_recommender_2016}.

\section{Methods}

\pop{Collaborative filtering method using mimic-iii.}A neighbourhood-based collaborative filtering method was used to model the recommender system, where similar bags, clustered using nearest neighbours algorithm, are used to make recommendations of tests for each other similar bag of laboratory tests. The data used to train and validate this model is the public demo data of the MIMIC-III dataset\footnote{MIMIC-III Clinical Database Demo -  \href{https://physionet.org/content/mimiciii-demo/1.4/}{https://physionet.org/content/mimiciii-demo/1.4/}}, even though I have access to the complete dataset, the demo data was selected for easy reproducibility.

\pop{Python package based on scikit-learn.}A python package named LaboRecommender was developed alongside this paper. This python package is completely based and compliant with the scikit-learn API. Every experiment communicated in this paper can be reproduced using the LaboRecommender package.

% \subsection{Data}

\pop{Mimic-iii description.}Medical Information Mart for Intensive Care III (MIMIC-III) is a database containing comprehensive clinical data relating to tens of thousands of Intensive Care Unit (ICU) patients. This dataset was used to extract test orders from real-world clinical encounters.

\pop{Mimic-iii labevents.}The table used for this paper was the \texttt{LABEVENTS}\footnote{MIMIC-III LABEVENTS Table -  \href{https://mimic.physionet.org/mimictables/labevents/}{https://mimic.physionet.org/mimictables/labevents/}} table which contains all laboratory measurements for a given patient. The attributes extracted from the table were specifically the name of the laboratory test and the order date and time. 
The demo dataset of the MIMIC-III dataset contains data from 100 different patients.

% \subsection{Preprocessing}

\pop{Labevents ordered at the same time are considered a bag.}Each laboratory test in the \texttt{LABEVENTS} table of MIMIC-III is a row, to extract the bags of laboratory tests, tests ordered at the same time were considered as ordered in a single bag. The number of different bags of laboratory tests extracted was 1596.

\pop{Bags are represented in a bag-item matrix.}To represent the bags used in this study and its contents, a bag-item matrix was used. This representation is a unary $m \times n$ matrix $V = [v_{uj}]$ containing $m$ bags and $n$ items. This matrix represent each bag as a row where $v_{uj} = 1$ if item $j$ is present in bag $u$.

% \subsection{Modeling}

\pop{Nearest neighbours.}To find the most similar bags to a query, a nearest neighbour search algorithm was used, where the \textit{distance function} is a hyperparameter. The functions we compared were Jaccard, Kulinski, Matching, Rogerstanimoto and Rusellrao, these functions were selected because they are custom tailored for unary vectors.

\pop{After selecting the neighbour bags, individual items are selected.}After selecting the $s$ most similar bags to the query (where $s$ is a hyperparameter) the frequency of each item in the retrieved bags is calculated to further sort the items in descending order respect to its frequency. Finally, I select the top $k$ items to return these items as the recommendation, this parameter needs to be selected after training depending on the implementation needs, 3, 5 and 10 recommendations were used to compute the metrics.

% \subsection{Validation}

\pop{Train test splitting, test subset, grid-search and cross-validation.}One-third of the dataset was selected for validation and the other two-thirds were used for training the model. For the optimization of the model hyperparameters, a grid-search with 5-fold cross-validation was used.

\pop{Mapk, mark maf1k.}Precision and recall metrics were used to validate the model over the testing subset implemented as a Mean average metric, namely Mean Average Precision (MAP) and Mean Average Recall (MAR). Mean average metrics are used to measure the performance of information retrieval systems which return a list of ranked results. For more information refer to \cite{manning_introduction_2008}.

\section{Results}

\pop{Grid search result.}The mean values of the 5-fold cross-validated precisions of each hyperparameter combination of the parameter grid are displayed in Table \ref{table:grid_search_result}. The best hyperparameter combination of the model is $s=20$ and Jaccard distance metric.

\pop{Metrics by n.}The performance metrics over the test subset of the model trained with the best hyperparameters are showed in Table \ref{table:performance_by_k}, these results are disaggregated by the number of tests suggested by the model. The value of $k=3$ achieved the best mean average precision and the value of $k=10$ achieved the best mean average recall.

\pop{Python package.}A fully documented Python package named LaboRecommender was developed to implement the algorithm proposed in this paper.

\begin{table}[!h]
\small
\centering
\begin{tabular}{lrrrrr}
\toprule
 & \multicolumn{5}{c}{Value of $s$} \\
Distance metric &    10  &    20  &    50  &    80  &    100 \\
\midrule
jaccard        & 93.67\% & \textbf{94.12\%} & 93.69\% & 93.46\% & 93.38\% \\
kulsinski      & 87.33\% & 89.69\% & 92.64\% & 93.17\% & 93.30\% \\
matching       & 84.48\% & 88.01\% & 89.06\% & 88.87\% & 89.37\% \\
rogerstanimoto & 84.78\% & 87.78\% & 89.03\% & 88.80\% & 89.33\% \\
russellrao     & 81.25\% & 85.96\% & 90.36\% & 92.21\% & 92.85\% \\
\bottomrule
\end{tabular}
\caption{Grid-search result. Values are mean 5-fold cross-validated precisions.}
\label{table:grid_search_result}
\end{table}
\vspace*{-\baselineskip}
\begin{table}[!h]
\small
    \centering
    \begin{tabular}{lrrr}
    \toprule
     & \multicolumn{3}{c}{Value of $k$} \\
    Performance metric &     3  &     5  &     10 \\
    \midrule
    MAP & \textbf{95.54\%} & 94.47\% & 91.83\% \\
    MAR & 18.51\% & 22.67\% & \textbf{31.13\%} \\
    \bottomrule
    \end{tabular}
    \caption{Recommender system performance metrics over test subset by $k$.}
    \label{table:performance_by_k}
\end{table}

\section{Discussion}

\pop{Best distance metric is jaccard and the value of s is 20.}Best hyperparameters were the usage of \textit{Jaccard distance} as the distance function between laboratory test sets and 20 as the number of most similar laboratory test sets in the neighbour search $s$. These hyperparameters achieved the best precision metric. These hyperparameters are consistent with current literature on systems that use nearest neighbour search as the core of its implementations \cite{niwattanakul_using_2013}.

\pop{Precision decreases with k and recall increases with k. Precision is better than recall.}Precision and recall behave inversely with the modulation of the parameter $k$ and for the application proposed in this paper the precision metric is the most important, because we want to retrieve the most number of relevant tests in each suggestion set to enhance the user experience, therefore I suggest a lower value of $k$.

\pop{Tradeoff between precision an usability.}To select the correct number for $k$, metrics must be taken into account but also user experience. Based on the results I recommend a $k$ of 5 to lower the time the user searches over the list of suggestions alongside the tradeoff between precision and recall.

\pop{Development of open source medical informatics is important.}To enhance the availability of evidence-based medical informatics software the code of the method described in this paper is open source and ready-available to deploy inside clinical decision support systems. In medical informatics, the release of software alongside the description of new methods should be a common practice.

% \pop{Similar recommender systems.}

% \pop{The usage of recommender systems saves resources and decrease errors.}

\section{Conclusion}
\pop{Collaborative filtering can be used for modeling.}Collaborative filtering algorithms can be used to develop a recommender system for laboratory tests. \pop{Clinical decision support system software can be developed and released.}This clinical decision support system can be packaged into a ready-to-use software released for free to the medical informatics community.

\printbibliography

\end{document}